# Title Page

**Title:** A Hybrid Swarm and Gravitation based feature selection algorithm for Handwritten *Indic* Script Classification problem

**Authors:** Ritam Guha[1], Manosij Ghosh[1], Pawan Kumar Singh[1], Ram Sarkar[1], Mita Nasipuri[1]

[1] Department of Computer Science and Engineering, Jadavpur University, Kolkata, India

**Email:** ritamguha16@gmail.com, manosij1996@gmail.com, pawansingh.ju@gmail.com, raamsarkar@gmail.com, mitanasipuri@gmail.com

**ORCID ids:**

Ritam Guha - 0000-0002-1375-777X

Manosij Ghosh- 0000-0003-2954-9876

Ram Sarkar - 0000-0001-8813-4086

**Abstract:** In any multi-script environment, handwritten script classification is of paramount importance before the document images are fed to their respective Optical Character Recognition (OCR) engines. Over the years, this complex pattern classification problem has been solved by researchers proposing various feature vectors mostly having large dimension, thereby increasing the computation complexity of the whole classification model. Feature Selection (FS) can serve as an intermediate step to reduce the size of the feature vectors by restricting them only to the essential and relevant features. In our paper, we have addressed this issue by introducing a new FS algorithm, called Hybrid Swarm and Gravitation based FS (HSGFS). This algorithm is made to run on 3 feature vectors introduced in the literature recently – Distance-Hough Transform (DHT), Histogram of Oriented Gradients (HOG) and Modified log-Gabor (MLG) filter Transform. Three state-of-the-art classifiers *namely*, Multi-Layer Perceptron (MLP), *k*-Nearest Neighbour (*k*-NN) and Support Vector Machine (SVM) are used for the handwritten script classification. Handwritten datasets, prepared at block, text-line and word level, consisting of officially recognized 12 *Indic* scripts are used for the evaluation of our method. An average improvement in the range of 2-5% is achieved in the classification accuracies by utilizing only about 75-80% of the original feature vectors on all three datasets. The proposed methodology also shows better performance when compared to some popularly used FS models.

# A Hybrid Swarm and Gravitation based feature selection algorithm for Handwritten *Indic* Script Classification problem

Ritam Guha, Manosij Ghosh, Pawan Kumar Singh, Ram Sarkar, Mita Nasipuri

Department of Computer Science and Engineering, Jadavpur University, Kolkata, India

{ritamguha16, manosij1996, pawansingh.ju, raamsarkar, mitanasipuri}@gmail.com

**Abstract:** In any multi-script environment, identification of script in which a particular document page is written has paramount importance before the same is fed to the respective Optical Character Recognition (OCR) engine for creating its machine editable version. Since last two decades, this complex pattern classification problem has been solved by researchers proposing various feature vectors mostly having large dimension, thereby increasing the computation complexity of the classification model. Feature Selection (FS) can serve as an intermediate step to reduce the size of the feature vectors by restricting them only to the essential and relevant features. In this paper, we have addressed this issue by introducing a new FS algorithm, called Hybrid Swarm and Gravitation based FS (HSGFS). This algorithm is made to run on 3 feature vectors applied recently for handwritten script identification – Distance-Hough Transform (DHT), Histogram of Oriented Gradients (HOG) and Modified log-Gabor (MLG) filter Transform. Three standard classifiers *namely*, Multi-Layer Perceptron (MLP), *k*-Nearest Neighbour (*k*-NN) and Support Vector Machine (SVM) are used for evaluating the performance of the optimal feature vectors selected by the proposed HSGFS. Handwritten datasets, prepared at block, text-line and word level, consisting of officially recognized 12 *Indic* scripts are used for the evaluation of our method. An average improvement in the range of 2-5% is achieved in the classification accuracies by utilizing only about 75-80% of the original feature vectors on all three datasets. The proposed methodology also shows better performance when compared to some popularly used FS models.

## 1. Introduction

The past decade has witnessed an increased availability of digital images and high capacity low cost storing devices. This has made storage of handwritten or printed documents in digital format a lot easier and budget-friendly. These handwritten documents are non-editable in nature. In order to achieve easy editing, maintenance, indexing, retrieval and transfer of contents, researchers throughout the world strive to develop various Optical Character Recognizer (OCR) which are currently used to convert images of handwritten, typed or printed text into machine readable text. This machine readable text produced by OCR engines are easily editable and maintainable but the problem with these OCRs is that they are largely script-specific (the writing style or graphical form of a language is known as script). This has not been an issue till there are only single-script documents but as document storage became a regular practice among a large group of people, it became a necessity to store and process the multi-script documents as well. As OCRs are script-specific, multi-script documents require multiple OCRs to be converted into machine readable configuration. The problem of converting multi-script documents into machine readable format can be solved by introducing a new layer of workflow ahead of OCR feeding of documents which is known as Automatic Script Classification (ASC). The entire workflow for an automated multi-script document storage system is shown in Fig. 1.

**Fig 1: Schematic representation of the multi-script document storage system.**

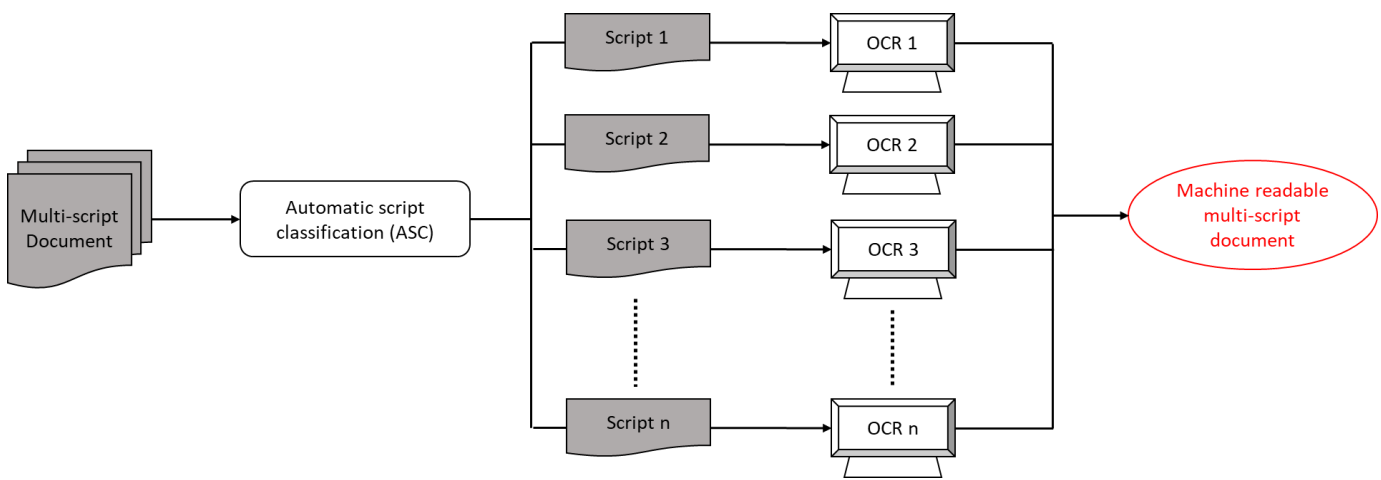

The difficulty of ASC can be easily realized by providing an overview of the vast set of languages that are currently used across the globe. According to Ethnologue catalogue of world languages, one of the best linguistic resources, currently there are 7,097 living languages used throughout the world [1]. A large multitude of scripts are constantly being used while these languages are expressed through writings. The presence of these large multitude of scripts has made the process of ASC very complex. The problem is profound in highly multilingual countries. One of the best examples of such a country is India which has 23 officially recognized languages and around 150 other languages. So, script/language identification has huge importance to increase digital communication in the field of culture, research and language studies. The 23 constitutionally recognized languages in India are *Bangla*, *English*, *Hindi*, *Punjabi*, *Marathi*, *Gujarati*, *Sindhi*, *Oriya*, *Assamese*, *Malayalam*, *Urdu*, *Telugu*, *Sanskrit*, *Tamil*, *Kannada*, *Nepali*, *Kashmiri*, *Maithali*, *Manipuri*, *Konkani*, *Bodo*, *Santhali*, and *Dogari* [2]. A total of 12 official scripts used to write the said Indian languages are: *Devanagari, Gujarati, Gurumukhi, Bangla, Tamil, Kannada, Urdu, Telugu, Malayalam, Manipuri, Oriya*, and *Roman*.

The identification of handwritten or printed scripts is a complicated process with a number of steps which includes pre-processing, segmentation, feature extraction and finally classification. Depending on the mode of segmentation, the problem of script classification can be conceived at three different levels: (1) word-level, (2) text line-level and (3) block-level. The next step is the creation of a feature vector through some feature extraction strategies. Many algorithms have been employed to extract features from the pre-processed and segmented documents. Each algorithm gives different feature vector which is then passed through a classifier for ASC. Sometimes the size of the feature vectors used for the ASC process becomes significantly large. Most of the times, such large feature vectors contain many redundant information which may decrease the overall classification accuracy provided by the classifiers. Even processing such large feature vectors results into huge time requirement. That is why before recognition, a FS algorithm can be used to keep the necessary and significant features which provide two advantages. Firstly, it extracts a near optimal set of features which improves the overall classification accuracy and secondly use of a reduced set of features decreases the load on the classifiers, thereby speeding up the recognition process. This fact motivates us to apply FS in the field of ASC for handwritten *Indic* scripts. There are mainly three categories of FS algorithms: filter [3,4], wrapper [5–7] and embedded [8–10]. Filter methods use the statistical measures of various features to select the optimal feature subset whereas wrapper methods take help of a classifier to check the classification capability of a feature subset. Due to supervision of a classifier, wrapper methods usually require more time than filter methods but on the other hand wrapper methods are able to perform better classification than filter methods. Now-a-days some of the researchers have proposed a hybrid version of both filter and wrapper versions which are known as embedded methods. For ASC, classification accuracy is lot more important than time requirement. Hence, we have focused on developing a hybridization of two popularly used wrapper models.

In this paper, we have proposed a hybridized version of Binary Particle Swarm Optimization (BPSO) and Binary Gravitational Search Algorithm (BGSA) known as Hybrid Swarm and Gravitation based FS (HSGFS) in order to get an optimal feature subset to be used for ASC from handwritten *Indic* script documents. Ebelhart and Kennedy in 1995 [6] created PSO which simulates the social behaviour observed in flocks of fish and birds. Over the years, PSO has gained huge popularity as a FS algorithm. In 2008, Rashedi and Saryazdi proposed GSA [7] to optimize solutions for single objective function. Contrasting to PSO, GSA works on the principles of Gravitational forces of Newtonian laws. PSO and GSA are two popularly used optimization algorithms but in our proposed model we have used their hybrid binary version to overcome the drawbacks of both the algorithms. A local search is also implemented within this hybrid version to improve the exploitation ability of the algorithm which is very useful to circumvent local optimal solution and reach the global optimal one. After getting the optimized output, the feature vector selected by HSGFS method is finally used to identify the script. The step-by-step procedure of handwritten ASC is represented in Fig. 2Fig . From the schematic representation presented in the figure, it can be observed that we have introduced a FS section (highlighted in red) to the existing workflow for ASC.

**Fig 2: Schematic representation of the handwritten ASC system using FS procedure.**

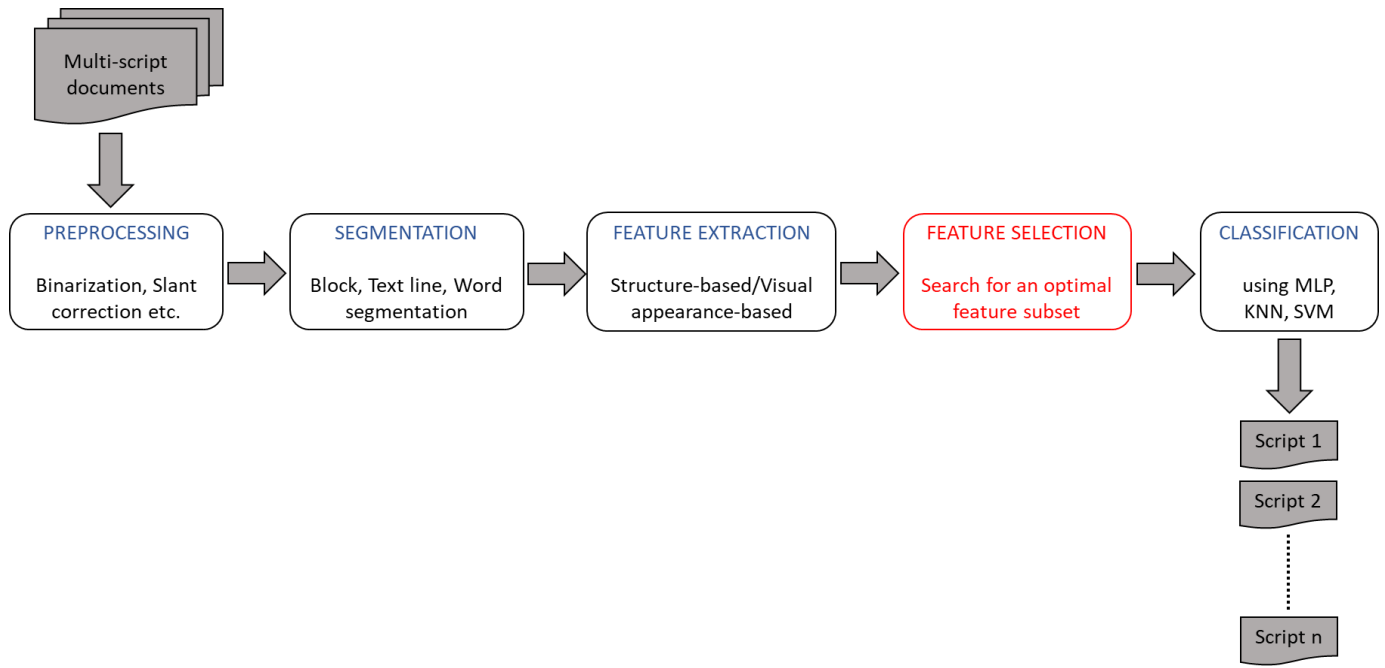

Our main contributions are enlisted below:

- Introducing a FS technique to the existing workflow of ASC to remove redundant and unimportant features from the extracted feature vectors thereby reducing the time requirement and improving the overall classification accuracy.

- Developing a new FS technique called HSGFS which combines the benefits of two existing FS models namely BPSO and BGSA. A local search technique is also incorporated in the hybrid model to improve the exploitation of the system of solutions.

- Applying the proposed FS model over handwritten datasets written in 12 *Indic* scripts at all the three different levels of ASC namely, block-level, text line-level and word-level.

- Comparing our proposed HSGFS procedure with existing FS models for handwritten ASC problem.

The rest of the paper is organized in 4 sections. Section 2 provides a brief description about some of the existing research works related to the domain of our work. The detailed explanation of our proposed FS model in ASC is provided in section 3. Section 4 consists of descriptions of the various experimentations we have performed to test and analyze the proposed method. Finally, section 5 concludes our work and provides a brief overview of the future scope mentioning the possible extension of this work.

## 2. Related Study

This section contains brief discussion of some previously proposed methods in the domain of script identification and FS. At first, we have discussed various script classification techniques and the later part of this section contains description of some significant variants of PSO and GSA found in the literature.

In 2017, Singh et al. [11] developed some standard datasets of handwritten *Bangla-Roman* and *Devanagari-Roman* mixed-script document images. Modified log-Gabor filter (MLG) was used for feature extraction in order to develop bi-script (*Devanagari-Roman* and *Bangla-Roman*) and tri-script (*Bangla-Devanagari-Roman*) word-level script identification modules.

In 2017, Obaidullah et al. [12] presented a handwritten document image dataset at page-level named *PHDIndic_11* having 11 officially recognized Indic scripts: *Devanagari, Bangla*, *Urdu, Roman*, *Oriya*, *Gujarati*, *Gurumukhi*, *Tamil*, *Malayalam, Telugu* and *Kannada*. The paper also contained the results for handwritten script identification (HSI). The authors used SL (Simple Logistics) and Multi-layer Perceptron (MLP) and their voting-based integration using average of probabilities in order to perform HSI.

In 2015, Singh *et al.* [13] performed script identification at word-level on some handwritten *Indic* scripts such as- *Devanagari*, *Bangla*, *Gurumukhi*, *Oriya*, *Malayalam*, *Telugu* and *Roman*. Feature extraction was done by a combination of *polygonal* and *elliptical approximation* and MLP was found to be the best classifier for ASC. In 2016, Chaudhari *et al.* [14] performed script classification of *Gujarati and English* languages at word-level. In order to perform feature extraction, the directional energy distribution of a word was obtained employing Gabor filters having suitable frequency and orientations. Their proposed model used SVM classifier to perform classification. In 2017, Obaidullah *et al.* [15] analyzed the performance of ASC when input data were conceived at different levels, i.e. page, block, text-line/word-level. The same multi-script handwritten document images were considered at 4 different levels and mainly 2 kinds of features were considered, *namely,* Script-Independent Features (SIF) and Script-Dependent Features (SDF). Final classification was performed by MLP and Random Forest (RF) classifiers. In 2017, Goswami *et al.* [16] put forward a novel approach for separating *Indic* scripts based on the presence of 'Matra', which was used as precursor to simplify following HSI in multi-script environment. Two different scripts, *Devanagari* and *Bangla*, were considered as positive samples as 'Matra' is present there; and two other scripts,

*Roman* and *Urdu*, were considered as negative samples for the experimentation as they do not have 'Matra'. After experimentation, Fractal Geometry Analysis (FGA) was found to be the best suited feature extraction methodology for script identification and RF classifier was the most appropriate classifiers among the classifiers used in the process. Singh et al. proposed a tree-oriented approach to perform recognition of 12 handwritten *Indic* scripts in [17]. The authors separated the Matra and non-Matra based scripts using Distance-Hough transform (DHT) and then they identified each script individually using modified log-Gabor (MLG) based features.

In 2018, Mukhopadhyay *et al.* proposed a method [18] to combine classifiers in order to efficiently recognize scripts in multilingual environment. Combination of classifiers reduced the complementary nature imposed by different classifiers on the same pattern. This reduced the burden of selecting appropriate classifier for a particular pattern recognition problem. The classifier combination approach was applied to handwritten *Indic* script (word-level) database developed by the authors which was named as *CMATERdb8.4.1* and was made online.

In 2015, Singh *et al.* [2] provided a survey on the feature extraction and script identification techniques used for the classification of printed or handwritten *Indic* scripts. The survey gave a platform to encourage future research activities in the field of script classification.

In spite of its importance, till date FS in script classification has been least explored. A FS approach for script identification was first attempted in printed documents using ReliefF algorithm[19]. In 2016, Das *et al.* [13] proposed a Harmony Search (HS) based FS procedure which was applied for HSI. In this approach, each candidate solution was considered as a musician. Just as musicians play various notes with different instruments and eventually find a perfect combination to get a harmony among the musical instruments, the candidate solutions were also fine-tuned and processed to achieve the most appropriate combination of frequencies (i.e. the final solution) which was then used to optimize the objective function. Thus, it can be seen that although FS can play an important role in ASC, yet it has not been attempted much by the researchers till now.

Throughout the years many optimization algorithms have been proposed which can be applied in the domain of FS. In 1995, Eberhart *et al.* [6] proposed two initial versions of PSO: GBEST (Global BEST) model and LBEST (Local BEST) model. GBEST model's candidate solutions (particles in case of PSO) use their own information as well as the information provided by the global best candidate to form their own solutions. Similarly, LBEST model particles produce solutions using their own information but instead of global best particle, they seek additional information from certain number of their neighbours. Canonical PSO utilises information obtained from only one neighbour, whereas in Full Information PSO (FIPSO) each neighbour acts as a source of information. Thus, Canonical PSO overlooks information provided by all the neighbours except one and FIPSO retains a lot of redundant features due to inclusion of too much information. To get rid of these limitations, Du et al. proposed an adequate-information version of PSO [21] in 2015 which was known as Limited Information PSO (LIPSO). LIPSO particles are influenced by top individuals of the swarm and the number of individuals influencing each particle may vary from particle to particle. In 2014, Cheng et al. incorporated social learning in PSO particles [22]. According to this approach, any particle may learn or retrieve information from the particles which are better than the individual of consideration. In 2007, Ghamisi *et al.* proposed a hybridized model combining Genetic Algorithm (GA) and Particle Swarm Optimization (PSO) [23]. The authors introduced elitism among the particles in the swarm. The elites or top performing particles in swarm qualifies to reach the next generation after going through PSO. All the other particles are discarded and GA is performed on the elites to obtain other candidates for the next generation.

In 2009, Rashedi *et al.* presented an optimization approach [7] based on Newtonian Laws of Motions and interaction of masses. The search agents were considered to be collections of masses interacting among themselves obeying laws of motions which guided the agents towards optimal solutions. A binary version of the GSA or BGSA has also been developed [24] by the same authors, Rashedi and Saryazdi to solve FS problems [25]. BGSA has been merged with Simulated Annealing(SA)to create GABSA [26]. Use of SA in GSA increases the local search and hence improves its exploitation abilities.

Recently Ghosh *et al.* proposed an improved version of GA known as Histogram based Multi-Objective GA (HMOGA) in [27]. They have produced optimized feature vectors for multiple runs of GA and combined the results using histogram-based cut-off criteria. In [28], Guha et al. proposed another updated version of GA where they replaced the mutation operation of GA with Great Deluge Algorithm (DGA) to improve the local searching capabilities of conventional GA. Like GA, Ant Colony Optimization (ACO) has gained massive popularity over the years as an efficient FS algorithm. Ghosh *et al.* introduced a wrapper-filter embedded version of ACO in [9]. The proposed model known as Wrapper-Filter ACOFS (WFACOFS) used both wrapper and filter methods to evaluate its candidate solutions thereby reducing the time requirement of the overall model (wrapper methods require more time to evaluate candidates than filter). Guha *et al.* proposed another level of improvement over HMOGA in

[29] where they added memory to the existing technique to store the best candidate solutions generated over the iterations which are eventually lost in the process. The model was then applied on handwritten numeral classification datasets.

Therefore, from the literature review, it can be noticed that application of FS in handwritten digit or word recognition is quite well-addressed. For example, in [29,30], the authors have applied FS for solving the problem of handwritten *Devanagari* digit recognition whereas the work described in [31] implements FS for handwritten *Bangla* word classification. But, to the best of our knowledge, FS has rarely been used for handwritten ASC problem. This motivates us to abridge this research gap and propose a FS method for handwritten ASC problem. In this work, the proposed FS method is applied on three state-of-the-art feature extraction algorithms namely, DHT algorithm [32], Histogram of Oriented Gradients (HOG) [33] and MLG Transform [34]. The proposed FS model is described in the next subsection.

## 3. Proposed Model

BPSO and BGSA are two well-known algorithms in the domain of FS. The first one is impressive in its exploitation abilities and the second one is rich in exploration. In order to attain a good exploitation-exploration trade-off, we have combined the methodologies of these two algorithms to create the new optimization method known as HSGFS. Even after combining these two algorithms, there remains a chance of premature convergence as both these algorithms tend to follow the global best in each iteration. Hence, we have implemented a local search within the algorithm to escape this convergence.

**3.1 Hybrid Swarm and Gravitation based Feature Selection (HSGFS)**

BPSOBGSA [35] is built combining the properties of both BPSO ($gbest$ which signifies the social thinking) and BGSA (which brings about its significant exploration abilities). The same concept is utilised in FS by converting the value of velocity to probability of whether a feature should be selected or not. A binary string ($B_j$) of size $n$ is used to signify the feature selection status. A '1' and a '0' denotes that the feature is selected and not selected respectively.

$$B_j = (f_{j1}, \dots, f_{jd}, \dots, f_{jn}), \qquad j = 1,2,\dots,N \qquad (1)$$

N is the total number of particles, $d$ is the index of the feature which is considered and $n$ is the number of features, and $f_{jd}$ is a binary value determining if the $jth$ feature is selected or not. Each particle in the population is randomly initialised with '0' and '1's.

Each particle influences the others with its influence proportional to its own fitness. The fitness of a particle is evaluated from the classification ability of the features chosen by that particle. The classification ability refers to the recognition accuracy of the candidate solutions (particles) obtained with the help of a trained classifier. The least performing particle ($worst(t)$) and the best performing particle ($best(t)$) are utilised to derive the masses of the particles using Equation 2. The masses are then modified using Equation 3 so as to allow the masses of each particle to be proportional to its relative strength. $fit_j(t)$ is the fitness (recognition accuracy) of agent $j$ at time $t$.

$$m_j = \frac{fit_j(t) - worst(t)}{best(t) - worst(t)} \quad (2)$$

$$M_j = \frac{m_j(t)}{\sum_{i=1}^{N} m_i(t)} \quad (3)$$

$G$, Gravitational constant, and $R_{ji}$, the Hamming distance between two particles $j$ and $i$, are obtained as follows:

$$G(t) = G_o * exp\left(-\alpha * \frac{iter}{max_{iter}}\right) \quad (4)$$

$$R_{ji} = \sqrt[2]{\sum_{k=0}^{D}(X_{jk}(t) - X_{ik}(t))^2} \quad (5)$$

α is the descending coefficient taken as 20 here. G₀ indicates the initial gravitational constant which is taken as 1, $iter$ denotes the current iteration number, and $max\_iter$ is the total number of iterations we set. Each particle asserts a force on another particle following Equation 6.

$$F_{ji}^d = G(t) * \left(\frac{M_j(t) * M_i(t)}{R_{ji}(t) + \varepsilon}\right) * \left(x_{id}(t) - x_{jd}(t)\right) \quad (6)$$

$M_j(t)$ is the mass related to agent $j$ at time $t$, $M_i$ is the mass corresponding to agent $i$, $G(t)$ is a gravitational constant at time instant $t$, $\varepsilon$ is a very small positive value, and $R_{ji}(t)$ is the Euclidean distance between two agents $j$ and $i$ at time instant $t$. $F_{ji}^d$ is the force between the two particles for the $dth$ feature. The total force on a particle is calculated by Equation 7 for all the $N$ particles. $r$ is a randomly generated number in the interval [0, 1].

$$F_j^d(t) = \sum_{i=1, i \neq j}^{N} r * F_{ji}^d(t) \qquad (7)$$

In accordance with the laws of motion, the value of acceleration is found by dividing the force by the mass of the particle (Equation 8). The force derived from the value of mass and the Hamming distance allow the particles to be influenced more by better particles which are similar (less Hamming distance). This value is then used to modify the velocity corresponding to the particle as shown in Equation 9. It combines the velocity updating strategies of both GSA and PSO. $c_1$ is the accelerating factor (Equation 10) and $c_2$ is the velocity factor (Equation 11).

$$a_{jd}(t) = \frac{F_{jd}(t)}{M_{jj}(t)} \qquad (8)$$

$$V_j(t+1) = r * V_j(t) + c_1 * a_j(t) + c_2 * \left(gbest - X_j(t)\right) \qquad (9)$$

$$c_1 = \left(-2 * \left(\frac{iter}{max_{iter}}\right)^3\right) + 2 \qquad (10)$$

$$c_2 = \left(2 * \left(\frac{iter}{max_{iter}}\right)^3\right) \qquad (11)$$

The value of velocity is interpreted as the probability of the feature being selected. The overall flow of the algorithm is represented in Fig. 3.

**Fig 3:** Flowchart of BPSO-BGSA.

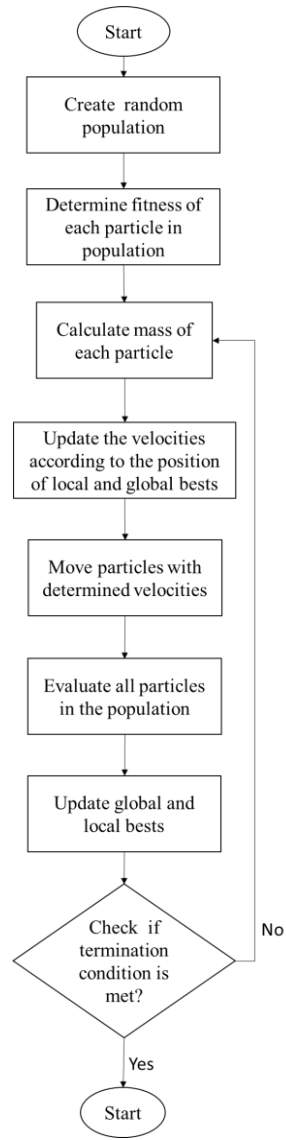

The problem with BPSO-BGSA as developed till now is the relatively poor local search capability especially in GPS which is pointed out in [36]. To account for this shortcoming, we introduce a local search method, where we perform a filter ranking in offline mode and utilise it in local search. In doing so, we add the top $k$ ranked features into a particle and delete the lowest $d$ ranked chromosomes we find in the particle. The ordered pair $(k, d)$ is generated randomly and lies between 1 and $floor((5*n)/100)$.

Another major problem of BPSO-BGSA is the lack of storage of the best feature sets. Therefore, if the population degrades to a worse solution over the iterations, the better solutions are lost. To account for this short coming, a memory has been added to the algorithm to retain the best solutions produced over the iterations. The overall steps of the proposed HSGFS are stated below:

| Step 1: | *Create a random set of particles (consisting of 0 and 1s).* |
| --- | --- |
| Step 2: | *Evaluate the fitness of the particles.* |
| Step 3: | *Recalculate the masses of the particles.* |
| Step 4: | *Move the particles following the laws of gravitation.* |
| Step 5: | *Revaluate the fitness of the particles.* |
| Step 6: | *Update the memory with the new particles.* |
| Step 7: | *Check if the stopping criteria are met.* |
| Step 8: | *If no then go to step 3.* |
| Step 9: | *Find the particle with highest classification accuracy and label it as final result.* |

### 3.2 Feature Extraction

The proposed FS method has been applied on three previously used feature descriptors such as DHT Algorithm [32], HOG [33] and MLG Transform [34] used for ASC. Since these feature descriptors have already been proposed earlier, so, here, these descriptors are described in brief.

### 3.2.1 Distance Hough Transform (DHT) Algorithm:

A feature vector consisting of 144 (72+72) attributes has been extracted using DHT algorithm. The steps of implementation of the DHT algorithm, as proposed in[32], are given below:

| Step 1: | Convert the RGB script image into a gray scale image. |
| --- | --- |
| Step 2: | Implement image smoothing operation using Gaussian low-pass filter [25]. |
| Step 3: | Implement image binarisation operation with a threshold value $\alpha$. The optimal value of $\alpha$ is found to be 0.7. |
| Step 4: | Invert the binarised image so that value '1' indicates foreground pixel and value '0' indicates background pixel. |
| Step 5: | Implement image thinning operation. |
| Step 6: | Implement Hough transform along 36 different orientations $\theta = -90°, -85°, -80°, \ldots, 0° \ldots, 75°, 80°$), with $\rho$ resolution taken as 1 pixel. |
| Step 7: | Calculate the $\rho$-value corresponding to the maximum accumulator value along each orientation. This generates 72 features. |
| Step 8: | Implement Euclidean Distance Transform [5] of the output word image from Step 5. |

**Step 9:** The Steps 6 and 7 are repeated to extract 72 features. Finally, the features produced from Steps 7 and 9 create a feature vector consisting of 144 elements for a specific script image.

**Step 10:** Exit.

### 3.2.2 Histogram of Oriented Gradients (HOG)

For object detection from images, the HOG descriptor was first proposed by Dalal and Triggs [33]. This descriptor was applied for detection of pedestrian from static images. The essential thought behind the HOG descriptors is that local object appearance and shape within an image can be described by the distribution of intensity gradients or edge directions. This method is similar to that of edge orientation histograms, scale-invariant feature transform descriptors, and shape contexts. The only difference is that it is computed on a dense grid of uniformly spaced cells and uses overlapping local contrast normalization for improved accuracy. The algorithm for implementing HOG descriptors is as follows:

**Step 1:** Convert the RGB script image into a gray scale image.

**Step 2:** Calculate the intensity gradient for each pixel. The intensity gradient is a vector with magnitude $m$ and orientation $\theta$ represented by the change in the intensity.

**Step 3:** Quantize the gradient orientations into 8 bins histogram for each cell (5x5 pixels) using the gradient magnitude and orientation.

**Step 4:** Compute a feature vector of 200 dimensions for each script image.

**Step 5:** Exit.

### 3.2.3 MLG Transform

MLG filter transform based features, as proposed in [34], is also considered as the one of texture feature descriptors for the classification of textual images based on the script in which it is written at three different levels *viz*., word-level, block-level and text line-level. In this work, a Windowed Fourier Transform (WFT) is taken into account for preserving the spatial information. The process of WFT consists of two steps. In the first step, the input image is multiplied with the window function whereas in the second step, the Fourier transform is applied to the previous step in order to get the resulting output. In short, WFT is mainly a convolution of the low-pass filter with the input image. MLG transform uses a Gaussian function as the optimally concerted function in both spatial as well as frequency domain [37].

In order to get the filtered images as output, the inverse Fourier transform is finally applied on the resulting vector. For the calculation of feature vector, two important measures such as energy and entropy features [38] are calculated from the MLG filter transformed images. Here, the number of scales ($n_s$) is chosen as 5 (that is, $n_s$=1, 2, 3, 4 and 5) and the number of orientations ($n_0$) is taken as 18 (that is, $10^0, 20^0, 30^0$, to $180^0$). Hence, this produces a feature set comprising 180 elements for a given input image containing handwritten text.

## 4. Experimental Outcomes and Analysis

This section describes the experimentation results achieved using the proposed FS method and related comparisons with other popularly used algorithms applied to ASC of handwritten *Indic* documents. All the experimentations of the proposed method are implemented in MATLAB 2016a environment and tested on PC with Intel Core-i3 (5$^{th}$ Gen.) CPU having 4 GB of RAM. The classification accuracy, used to measure the performance of handwritten ASC, is calculated as follows:

$$Classification\ Accuracy\ (\%) = \frac{\#\ successfully\ classified\ components}{\#total\ components\ present} \times 100 \quad (12)$$

### 4.1 Preparation of Handwritten *Indic* Script Database

We have prepared our own datasets for handwritten *Indic* script documents in the laboratory due to unavailability of the same in public domain. These document pages are collected from different writers who contributed their handwriting on A-4 sized white pages. We then scan the input script images at 300 dpi and save them in grey-scale form. Gaussian filter is then used to de-noise the noisy pixels in the collected images. Firstly, handwritten text blocks *written* in 12 *Indic* scripts, of pre-defined size 256x256 pixels are automatically cropped from the document pages. The extracted text blocks also have a chance of containing lines of varying size, thickness and white spaces between characters, lines or words. Instead of performing any homogenizing technique to compensate for this, we try to manually ensure that at least 50% of our input image region contains text. Fig. 4 shows samples of handwritten text blocks in 12 Indian scripts. In a similar manner, the words and then the test-lines are also extracted from the input document pages employing the techniques described in [39] and [40] respectively. Finally, a set of 7200 text blocks (with exactly 600 text blocks per script), 3600 text lines (with exactly 300 text-lines per script) and 12000 text words (with exactly 1000 text words per script) written in 12 *Indic* scripts are prepared to evaluate the proposed HSGFS methodology for ASC from handwritten text obtained at block, text line and word level respectively.

**Fig. 4: Samples of text blocks taken from our database written in 12 official scripts of India.**

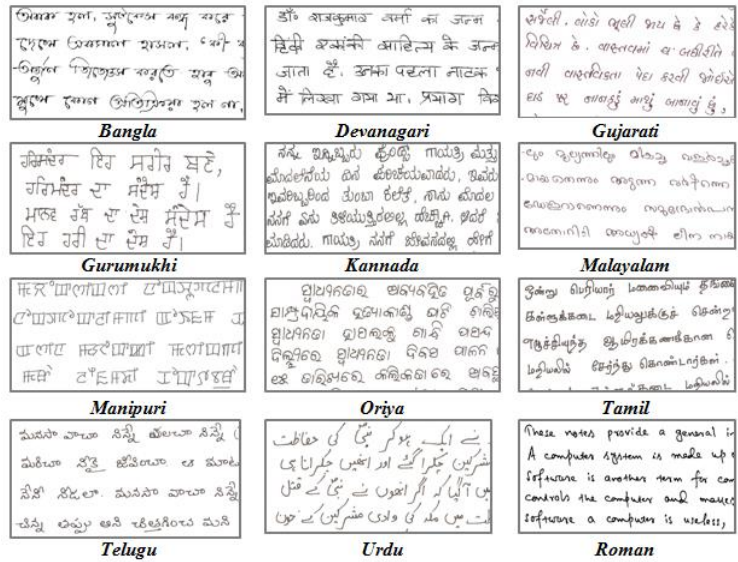

## 4.2 Parameter Setting of HSGFS methodology

The performances of DHT algorithm, HOG and MLG Transform feature sets is observed by altering the size of population and number of iterations, the two most important parameters of our FS method called HSGFS. This is done to select the optimal values for these two parameters in the present context. Throughout the process of searching for the optimal values of the parameters, MLP has been used as the classifier. After initial experimentation, it is observed that setting population size as 20 and number of iterations as 15 gives good results. Then we have varied one of these values keeping the other one constant. Firstly, the number of iterations is kept constant (in present case, 15) while the population size is varied for all the three datasets. Outcome of this experiment is graphically illustrated in Fig. 5. Then, the population size is kept constant (in our case, 20) while the number of iterations is varied which is depicted in Fig. 6. It can be observed from Fig. 5 (a-c) that the optimal values of population size are 15, 20 and 20 for DHT algorithm, HOG and MLG Transform feature sets respectively. From Fig. 6 (a-c) it is also clear from that the optimal number of iterations is 15 for all three feature sets.

**Fig 5: Performance comparison illustrating the variation of classification accuracy with population size keeping number of iterations fixed on block, text line and word-level datasets for: (a) DHT algorithm, (b) HOG and (c) MLG Transform feature sets.**

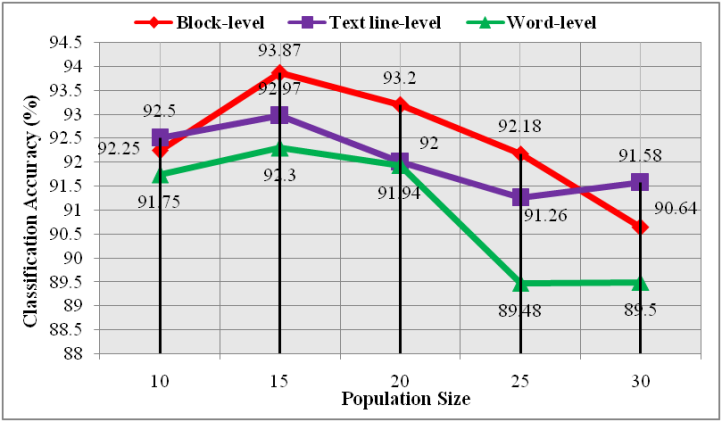

(a)

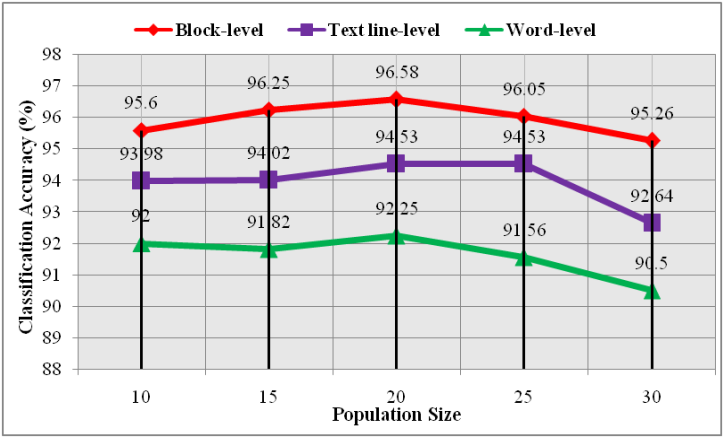

(b)

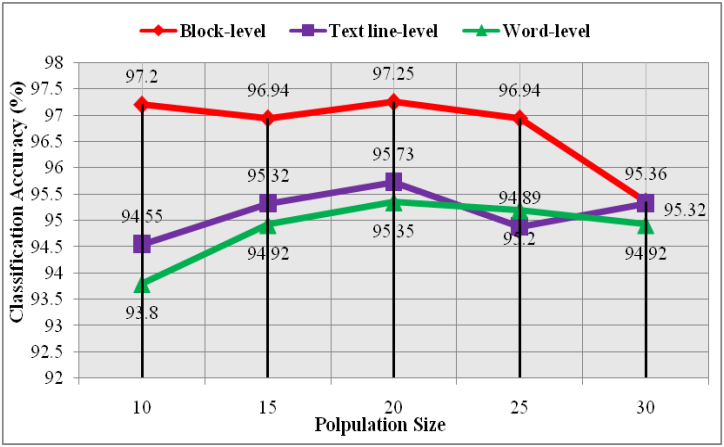

(c)

**Fig 6: Performance comparison illustrating the variation of classification accuracy with number of iterations keeping population size fixed on block, text line and word-level datasets for: (a) DHT algorithm, (b) HOG and (c) MLG Transform feature sets.**

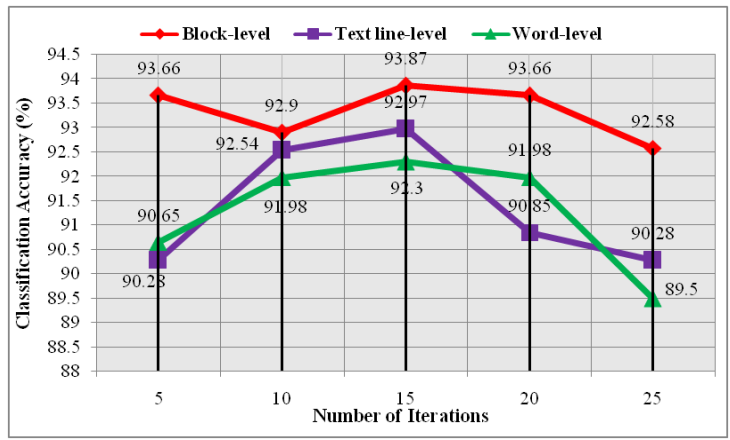

(a)

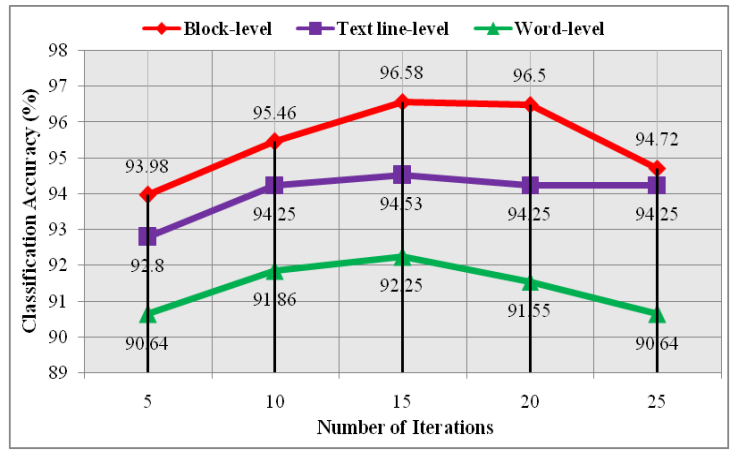

(b)

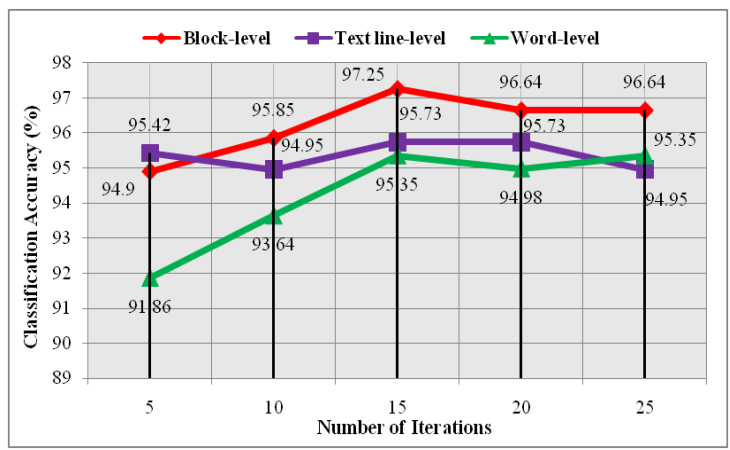

(c)

The results obtained from the variation of parameters indicate that our initial estimations of number of iterations being 15 and population size being 20 are optimal with minor variations. Hence, we have used these values for population size and number of iterations for rest of the experimentations.

**4.3 ASC performance results without using HSGFS methodology**

Firstly, previously mentioned three feature sets have been used separately for classification of the *Indic* scripts on all the three types of datasets at three different levels without any FS (i.e. all available features are used for identification). Here, training and testing of handwritten script samples are done using three popular classifiers *namely*, MLP [41], *k*-NN [42] and SVM [43]. The script classification accuracies attained by these three classifiers, without using any FS scheme, are noted in Table 1. It can be witnessed from Table 1 that in case of DHT algorithm, *k*-NN classifier scores the highest classification accuracies of 91.62%, 92.16% and 87.25% at block, text-line and word-level datasets respectively. In case of HOG feature descriptor, SVM classifier records the highest classification accuracies of 93.16%, 94.55% and 89.6% at block, text-line and word-level datasets respectively. Whereas, for MLG transform, the highest classification accuracies of 93.15%, 95.8% and 91.34% are attained by MLP classifier at block, text-line and word-level datasets respectively.

**Table 1 : ASC accuracies of three different original feature sets (without FS) using MLP, *k*-NN and SVM classifiers on block, text-line, and word-level datasets (maximum accuracy achieved at each level is marked in bold style)**

| Feature vector | Size of feature vector | Level of classification | Number of Training samples | Number of Testing samples | Classification accuracy without FS | | |
|---|---|---|---|---|---|---|---|
| | | | | | MLP classifier | *k*-NN classifier | SVM classifier |
| DHT Algorithm | 144 | Block-level | 400 | 200 | 90.56% | **91.62%** | 89.24% |
| | | Text line-level | 200 | 100 | 91.02% | **92.16%** | 91.39% |
| | | Word-level | 650 | 350 | 85.69% | **87.25%** | 84.40% |
| HOG | 200 | Block-level | 400 | 200 | 91.50% | 92.33% | **93.16%** |
| | | Text line-level | 200 | 100 | 92.26% | 91.62% | **94.55%** |
| | | Word-level | 650 | 350 | 88.41% | 88.94% | **89.60%** |

| | | Block-level | 400 | 200 | **93.15%** | 92.50% | 91.19% |
| MLG Transform | 180 | Text line-level | 200 | 100 | **95.80%** | 92.18% | 93.95% |
| | | Word-level | 650 | 350 | **91.34%** | 89.95% | 90.08% |

## 4.5 ASC performance results using HSGFS methodology

### 4.5.1 MLP Classifier

MLP classifier, described in [41], is used to measure the performance of the optimal feature sets produced by the proposed HSGFS methodology. The values of two parameters *namely*, $\eta$ and $\alpha$ for MLP classifier are experimentally set to be 0.6 and 0.5 respectively and the classifier is made to run for 1000 epochs. The MLP classification model is taken as $\alpha' - \beta' - \gamma'$. Here, $\gamma'$ and $\alpha'$ are defined as the number of neurons present in the output and input layers which are taken as the number of output classes and the number of features considered here respectively. The number of neurons present in hidden layer (denoted by $\beta'$) for the three feature sets is varied experimentally in order to achieve the optimal results. Appendix A contains the graphical representation of the various results obtained through alteration of number of neurons for MLP classifier. Then, we perform ASC on the previously mentioned datasets after performing FS using our proposed method HSGFS. Table 2 depicts that for DHT algorithm, the highest classification accuracies of 95.7%, 95.91% and 93.73% are achieved using MLP classifier on block, text line, and word-level datasets. The numbers of optimal features to attain the best accuracies are 114, 112 and 118. For HOG feature set, the highest classification accuracies of 97.33%, 96.98% and 94.7% are achieved using MLP classifier on block, text line, and word-level datasets. The numbers of optimal features selected for this case are 140, 149 and 164. Similarly, for MLG Transform feature set, the highest classification accuracies of 98.06%, 97.74% and 96.28% are achieved using MLP classifier on block, text line, and word-level datasets. The numbers of optimal features selected for this case are 132, 130 and 138. It can be clear that the MLG transform feature set performs the best among all the three feature sets when MLP classifier is used.

**Table 2 : ASC accuracies of three different feature sets after FS by HSGFS using MLP, $k$-NN and SVM classifiers on block, text line, and word-level datasets (maximum accuracy achieved at each level is made bold).**

| Feature vector | Level of classification | Classification accuracy after FS using HSGFS | | | | | |
|---|---|---|---|---|---|---|---|
| | | MLP classifier | | k-NN classifier | | SVM classifier | |
| | | Classification accuracy | Size of Feature Vector | Classification accuracy | Size of Feature Vector | Classification accuracy | Size of Feature Vector |
| DHT Algorithm | Block-level | **95.7%** | 114 | 94.75% | 115 | 94.95% | 114 |
| | Text line-level | **95.91%** | 112 | 93.62% | 118 | 94.75% | 117 |
| | Word-level | 93.73% | 118 | 93.13% | 120 | **93.97%** | 116 |
| HOG | Block-level | **97.33%** | 140 | 97.02% | 144 | 97.12% | 142 |
| | Text line-level | **96.98%** | 149 | 95.42% | 118 | 96.83% | 151 |
| | Word-level | **94.7%** | 164 | 93.05% | 149 | 93.97% | 168 |
| MLG Transform | Block-level | **98.06%** | 132 | 97.93% | 146 | 97.58% | 155 |
| | Text line-level | **97.74%** | 130 | 96.63% | 133 | 97.08% | 134 |
| | Word-level | 96.28% | 138 | 95.83% | 127 | **96.73%** | 137 |

### 4.5.2 *k*-NN Classifier

*k*-NN classifier [42] is next chosen for evaluation of the proposed HSGFS methodology and is made to run for different values of *k* on all the three types of script datasets. The experimental value of *k* is made to vary from 2 to 6. Appendix A contains the graphical representation of the various results obtained through alteration of k-value for *k*-NN classifier. Table 2 depicts that for DHT algorithm, the highest classification accuracies of 94.75%, 93.62% and 93.13% are achieved using *k*-NN classifier on block, text line, and word-level datasets. The numbers of optimal features selected required to attain the best accuracies are 115, 118 and 120. For HOG feature set, the maximum classification accuracies of 97.02%, 95.42% and 93.05% are achieved $k$-NN classifier on block, text line, and word-level datasets. The numbers of optimal features selected for this case are 144, 118 and 149. Similarly, for MLG transform feature set, the highest classification accuracies of 97.93%, 96.63% and 95.83% are achieved using $k$-NN classifier on block, text line, and word-level datasets. The numbers of optimal features selected for this case are 146, 133 and 127. It can be clear that the MLG transform feature set again performs the best among all the three feature sets for $k$-NN classifier and the optimal results are achieved for $k=3$.

### 4.5.3 SVM Classifier

Finally, SVM classifier with polynomial kernel is employed to measure the FS ability of the proposed HSGFS methodology on the three script datasets. Table 2 illustrates that for DHT algorithm, the highest classification accuracies of 94.95%, 94.75% and 93.97% are achieved using SVM classifier on block, text line, and word-level datasets. The numbers of optimal features required to attain the best accuracies are 114, 117 and 116. For HOG feature set, the highest classification accuracies of 97.12%, 96.83% and 93.97% are achieved using SVM classifier on block, text line, and word-level datasets. The numbers of optimal features selected for this case are 142, 151 and 168. Similarly, for MLG Transform feature set, the highest classification accuracies of 97.58%, 97.08% and 96.73% are achieved using SVM classifier on block, text line, and word-level datasets. The numbers of optimal features selected for this case are 155, 134 and 137. Moreover, MLG Transform feature set once again performs the best among all the three feature sets when SVM classifier is applied.

### 4.6 Summarization of performance results of HSGFS methodology

It is already clear from the above trial outcomes that the best feature descriptor is found to be MLG Transform as it showed the highest classification accuracies on all three script datasets. For block-level dataset, the best classification accuracy of 98.06% (considering only 132 features) is achieved by MLP classifier. For text line-level dataset, maximum accuracy of 97.74% (utilizing 130 features) is attained by MLP classifier whereas SVM classifier provides the best overall accuracy of 96.73% (considering only 137 features) on word-level dataset. The above results suggest that the script classification accuracy decreases as we move from block-level to word-level. Table 3 summarizes the overall results for block, text line and word-level datasets. It is evident from Table 3 that after applying the proposed HSGFS method, the sizes of optimal feature sets are found to be 132, 130 and 137 for block, text line and word-level datasets respectively. This means that the proposed HSGFS methodology selects only about

73%, 72% and 76% of the original feature vectors for the three datasets respectively. Moreover, increments of about 5%, 2% and 5% in the original classification accuracies are also noticed in case of the three datasets.

Table 3: Overall summarization of performance results of our proposed FS method, called HSGFS, on block, text line and word-level datasets.

| Datasets | Before FS | | | After FS | | | | |
|---|---|---|---|---|---|---|---|---|
| | Size of original feature set | Best classifier | Original classification accuracy | Size of optimal feature set | Best classifier | Reported best classification accuracy | Number of original features considered (in %) | Increment in accuracy |
| Block-level | 180 | SVM | 93.16% | 132 | MLP | 98.06% | 73.33 | 4.90% |
| Text line- | 180 | MLP | 95.80% | 130 | MLP | 97.74% | 72.22 | 1.94% |
| Word-level | 180 | MLP | 91.34% | 137 | SVM | 96.73% | 76.11 | 5.39% |

**4.7 Performance comparison of HSGFS methodology with other well-known optimization algorithms**

It can be seen that FS can increase the accuracy of the ASC system. But, there exist several previously proposed optimization algorithms which can be used to perform FS. So, in order to evaluate the efficiency of the present FS method i.e. HSGFS, we have provided a comparison of some popularly used FS methods such as GA, PSO, GSA, SA, HS with HSGFS when applied for solving ASC problem. We have used three classifiers for evaluation of our proposed method and MLP was found to provide the best classification accuracy. Hence, we have compared the FS performance of HSGFS with the other previously mentioned algorithms in terms of classification accuracies obtained by using MLP classifier. From Table 4, it can be seen that HSGFS outperforms other optimization algorithms at all the three different levels of ASC in terms of classification accuracy.

**Table 4 : Comparison of HSGFS with other well-known optimization algorithms when applied in ASC (the best accuracy obtained at each level is made bold)**

| Feature Vector | Level of classification | HSGFS (Proposed) | | PSO | | GSA | | GA | | SA | |
|---|---|---|---|---|---|---|---|---|---|---|---|
| | | Classification Accuracy | Optimal size of Feature set | Classification Accuracy | Optimal size of Feature set | Classification Accuracy | Optimal size of Feature set | Classification Accuracy | Optimal size of Feature set | Classification Accuracy | Optimal size of Feature set |
| DHT Algorithm | Block-level | **95.7%** | 114 | 60.41% | 79 | 60.75% | 71 | 64.62% | 69 | 66.12% | 81 |
| | Text line-level | **95.91%** | 112 | 83.83% | 83 | 85.16% | 100 | 90% | 96 | 83.16% | 68 |
| | Word-level | **93.73%** | 118 | 74.55% | 82 | 76.3% | 105 | 79.81% | 106 | 72.33% | 66 |
| HOG | Block-level | **97.33%** | 140 | 71.83% | 98 | 72.62% | 112 | 74.83% | 113 | 76.33% | 96 |
| | Text line-level | **96.98%** | 149 | 78.75% | 117 | 80.83% | 106 | 86.67% | 140 | 82.75% | 102 |
| | Word-level | **94.7%** | 164 | 78.33% | 119 | 80.78% | 141 | 85.11% | 127 | 82.81% | 111 |
| MLG | Block-level | **98.06%** | 132 | 67.54% | 107 | 68.91% | 90 | 73.58% | 122 | 76.08% | 91 |
| | Text line-level | **97.74%** | 130 | 95.83% | 109 | 96.04% | 98 | 94.06% | 116 | 93.83% | 95 |
| | Word-level | **96.28%** | 138 | 92.5% | 100 | 92.86% | 126 | 94.33% | 110 | 93.33% | 92 |

### 4.8 Comparison with previously proposed FS techniques reported for ASC

Since, there is hardly any work found in the literature which uses FS for handwritten ASC problem, in the present work, we have compared the proposed HSGFS with one of the previously proposed works where Harmony Search (HS) based FS technique is reported [13]. The same MLG Transform feature vector is implemented on the current datasets and classified using MLP classifier. It can be noticed from comparative analysis given in Table 5 that using lesser number of features, the present HSGFS methodology performs better than the HS based FS method for all the three levels of ASC.

**Table 5: Performance comparison of the present HSGFS method with the method described in [13]**

|  |  | HS based FS methodology proposed by Singh *et al.* [13] | Present HSGFS Methodology |
|---|---|---|---|
| **Size of original feature vector** |  | 180 | 180 |
| **Size of optimal feature vector** | **Block-level** | 156 | 132 |
|  | **Text line-level** | 139 | 130 |
|  | **Word-level** | 155 | 137 |
| **Classification Accuracy (%)** | **Block-level** | 94.02 | 98.06 |
|  | **Text line-level** | 96.39 | 97.74 |
|  | **Word-level** | 91.6 | 96.73 |

### 5. Conclusion

Though FS is an important research topic in the area of ASC, researchers have not yet addressed it in the domain of handwritten *Indic* script identification. In this work, a new hybridized version of BPSO and BGSA, called HSGFS, has been proposed for implementing FS. The FS capability of the proposed method has been tested on three different feature sets *namely*, MLG, HOG and DHT at three different levels of script classification. It can be witnessed from the experimental results that the proposed HSGFS method attains increments of around 5%, 2% and 5% with respect to when no FS method has been applied, and choosing only around 73%, 72% and 76% of the original feature vectors for the block, text-line and word-level datasets respectively. Our proposed HSGFS method is also found to perform better than other renowned optimization algorithms like GSA, PSO, GA, SA and HS. As a future scope, we can suggest to increase the number of datasets on which HSGFS is applied. Some other scripts apart from *Indic* scripts can be used to ascertain the applicability of the proposed technique. Moreover, as the proposed method is applicable to any pattern recognition problem, the proposed HSGFS model can be employed to solve other PR problems like facial emotion recognition, word identification, character recognition etc. to test its efficiency.

**Compliance with Ethical Standards:**

**Funding**: This study was self-funded.

**Conflict of Interest**: All the authors declare that they have no conflict of interest.

**Ethical approval**: This article does not contain any studies with human participants performed by any of the authors.

## References

[1] https://en.wikipedia.org/wiki/Ethnologue accessed on 20-05-2018, (n.d.). https://en.wikipedia.org/wiki/Ethnologue.

[2] P.K. Singh, R. Sarkar, M. Nasipuri, Offline Script Identification from multilingual Indic-script documents: A state-of-the-art, Comput. Sci. Rev. 15 (2015) 1–28. doi:10.1016/j.cosrev.2014.12.001.

[3] H. Liu, H. Motoda, Computational Methods of Feature Selection, CRC Press, 2007. doi:10.1201/9781584888796.

[4] P. Mitra, C.A. Murthy, S.K. Pal, Unsupervised feature selection using feature similarity, IEEE Trans. Pattern Anal. Mach. Intell. 24 (2002) 301–312.

[5] M. Dorigo, M. Birattari, Ant colony optimization, in: Encycl. Mach. Learn., Springer, 2011: pp. 36–39.

[6] R. Eberhart, J. Kennedy, A new optimizer using particle swarm theory, in: Micro Mach. Hum. Sci. 1995. MHS'95., Proc. Sixth Int. Symp., IEEE, 1995: pp. 39–43.

[7] E. Rashedi, H. Nezamabadi-pour, S. Saryazdi, GSA: A Gravitational Search Algorithm, Inf. Sci. (Ny). 179 (2009) 2232–2248. doi:10.1016/j.ins.2009.03.004.

[8] M. Ghosh, S. Begum, R. Sarkar, D. Chakraborty, U. Maulik, Recursive Memetic Algorithm for gene selection in microarray data, Expert Syst. Appl. 116 (2019) 172–185.

[9] M. Ghosh, R. Guha, R. Sarkar, A. Abraham, A wrapper-filter feature selection technique based on ant colony optimization, Neural Comput. Appl. (2019) 1–19. doi:https://doi.org/10.1007/s00521-019-04171-3.

[10] M. Ghosh, S. Adhikary, K.K. Ghosh, A. Sardar, S. Begum, R. Sarkar, Genetic algorithm based cancerous gene identification from microarray data using ensemble of filter methods, Med. Biol. Eng. Comput. 57 (2019) 159–176.

[11] P.K. Singh, R. Sarkar, N. Das, Benchmark databases of handwritten Bangla - Roman and Devanagari - Roman mixed-script document images, Multimedia Tools and Applications, 2018. doi:10.1007/s11042-017-4745-3.

[12] S.M. Obaidullah, C. Halder, K.C. Santosh, N. Das, K. Roy, PHDIndic_11: page-level handwritten document image dataset of 11 official Indic scripts for script identification, Multimed. Tools Appl. 77 (2018) 1643–1678. doi:10.1007/s11042-017-4373-y.

[13] P.K. Singh, S. Das, R. Sarkar, M. Nasipuri, Feature Selection Using Harmony Search for Script Identification from Handwritten Document Images, J. Intell. Syst. 27 (2018) 465–488.

[14] S. Chaudhari, M. Gulati, Script Identification using Gabor Feature and SVM Classifier, Procedia - Procedia Comput. Sci. 79 (2016) 85–92. doi:10.1016/j.procs.2016.03.012.

[15] S. Obaidullah, Automatic Indic script identification from handwritten documents : page , block , line and word - level approach, Int. J. Mach. Learn. Cybern. 0 (2017) 0. doi:10.1007/s13042-017-0702-8.

[16] C. Goswami, K.C. Santosh, N. Das, C. Halder, K. Roy, Separating Indic Scripts with matra for E ective Handwritten Script Identi cation in Multi-Script Documents, 31 (2017). doi:10.1142/S0218001417530032.

[17] P.K. Singh, R. Sarkar, V. Bhateja, M. Nasipuri, A comprehensive handwritten Indic script recognition system: a tree-based approach, J. Ambient Intell. Humaniz. Comput. (2018) 1–18.

[18] A. Mukhopadhyay, P.K. Singh, R. Sarkar, M. Nasipuri, A Study of Different Classifier Combination Approaches for Handwritten Indic Script Recognition, (2018). doi:10.3390/jimaging4020039.

[19] V. Ablavsky, M.R. Stevens, Automatic feature selection with applications to script identification of degraded documents, in: Null, IEEE, 2003: p. 750.

[20] et al. (2017). Singh, P., Das, S., Sarkar, R., Feature Selection Using Harmony Search for Script Identification from Handwritten Document Images. Journal of Intelligent Systems, 0(0), pp. -. Retrieved 19 Jun. 2018, from doi:10.1515/jisys-2016-0070, (n.d.).

[21] W. Du, Y. Gao, C. Liu, Z. Zheng, Z. Wang, Adequate is better : particle swarm optimization with, Appl. Math. Comput. 268 (2015) 832–838. doi:10.1016/j.amc.2015.06.062.

[22] R. Cheng, Y. Jin, A social learning particle swarm optimization algorithm for scalable optimization, Inf. Sci. (Ny). 291 (2015) 43–60. doi:10.1016/j.ins.2014.08.039.

[23] P. Ghamisi, S. Member, J.A. Benediktsson, Feature Selection Based on Hybridization of Genetic Algorithm and Particle Swarm Optimization, 12 (2015) 309–313. doi:10.1109/LGRS.2014.2337320.

[24] E. Rashedi, H. Nezamabadi-Pour, S. Saryazdi, BGSA: Binary gravitational search algorithm, Nat. Comput. 9 (2010) 727–745. doi:10.1007/s11047-009-9175-3.

[25] FEATURE SELECTION THROUGH GRAVITATIONAL SEARCH ALGORITHM Department of Computing University of S ̃ ao Paulo University of Campinas Institute of Computing, Sort. (2011) 2052–2055.

[26] C. Tong, Gravitational search algorithm based on simulated annealing, J. Converg. Inf. Technol. 9 (2014) 231.

[27] M. Ghosh, R. Guha, R. Mondal, P.K. Singh, R. Sarkar, Feature Selection using

Histogram based Multi-Objective GA for Handwritten Devanagari Numeral Recognition, (2017).

[28] R. Guha, M. Ghosh, S. Kapri, S. Shaw, S. Mutsuddi, V. Bhateja, R. Sarkar, Deluge based Genetic Algorithm for feature selection, Evol. Intell. (2019) 1–11.

[29] R. Guha, M. Ghosh, P.K. Singh, R. Sarkar, M. Nasipuri, M-HMOGA: A New Multi-Objective Feature Selection Algorithm for Handwritten Numeral Classification, J. Intell. Syst. (2019). doi:https://doi.org/10.1515/jisys-2019-0064.

[30] M. Ghosh, R. Guha, R. Mondal, P.K. Singh, R. Sarkar, M. Nasipuri, Feature selection using histogram-based multi-objective GA for handwritten Devanagari numeral recognition, 2018. doi:10.1007/978-981-10-7566-7_46.

[31] M. Ghosh, S. Malakar, S. Bhowmik, R. Sarkar, M. Nasipuri, Feature Selection for Handwritten Word Recognition Using Memetic Algorithm, in: Adv. Intell. Comput., Springer, 2019: pp. 103–124.

[32] P. K. Singh, S. Das, R. Sarkar, M. Nasipuri, "Line Parameter based Word-Level Indic Script Identification System", In: International Journal of Computer Vision and Image Processing, IGI Global Publishers, Vol. 6, Issue 2, pp. 18-41, July-December 2016., (n.d.).

[33] N. Dalal, B. Triggs, "Histograms of Oriented Gradients for Human Detection", http://lear.inrialpes.fr, (n.d.).

[34] P. K. Singh, I. Chatterjee, R. Sarkar, "Page level Handwritten Script Identification using Modified log-Gabor filter based features", In: Proc. Of 2nd IEEE International Conference on Recent Trends in Information Systems (ReTIS), pp. 225-230, Kolkata, Ind, (n.d.).

[35] S. Mirjalili, S.Z.M. Hashim, A new hybrid PSOGSA algorithm for function optimization, Proc. ICCIA 2010 - 2010 Int. Conf. Comput. Inf. Appl. (2010) 374–377. doi:10.1109/ICCIA.2010.6141614.

[36] Angeline PJ (1998) Evolutionary optimization versus particle swarm optimization: Philosophy and performance differences. In: International Conference on Evolutionary Programming. Springer, pp 601–610., (n.d.).

[37] J.G. Daugman, Uncertainty relation for resolution in space, spatial-frequency,

and orientation optimized by two-dimensional visual cortical filters., J. Opt. Soc. Amer. 2 (n.d.) 1160–1169.

[38] R. C. Gonzalez, R. E. Woods, "Digital Image Processing", Vol. I, Prentice-Hall, India (1992)., (n.d.).

[39] R. Sarkar, Pattern Recognition and Machine Intelligence, 2011. doi:10.1007/978-3-642-21786-9.

[40] P. K. Singh, S. P. Chowdhury, S. Sinha, S. Eum, R. Sarkar: "Page-to-Word Extraction from Unconstrained Handwritten Document Images", In: Proc. of 1st International Conference on Intelligent Computing and Communication (ICIC2), AISC 458, pp. 517-524, 2, (n.d.).

[41] S. Basu, N. Das, R. Sarkar, M. Kundu, M. Nasipuri, D.K. Basu, Handwritten ' Bangla ' alphabet recognition using an MLP based classfier, in: 2nd Natl. Conf. Comput. Process. Bangla-2005, 2005: pp. 285–291.

[42] P.E.H. T. M. Cover, Nearest neighbor pattern classification, In: , IT-13(1), pp. 21–27, 1967., IEEE Trans. Inform. Theory. 13 (1967) 21–27.

[43] L. Saitta, Support-Vector Networks, 297 (1995) 273–297.

**Appendix A**

**Fig. 1: Graphical comparison showing the variation of classification accuracy with the number of neurons in hidden layer of MLP classifier for: (a) DHT algorithm, (b) HOG and (c) MLG transform feature sets on block-level, text line-level, and word-level datasets.**

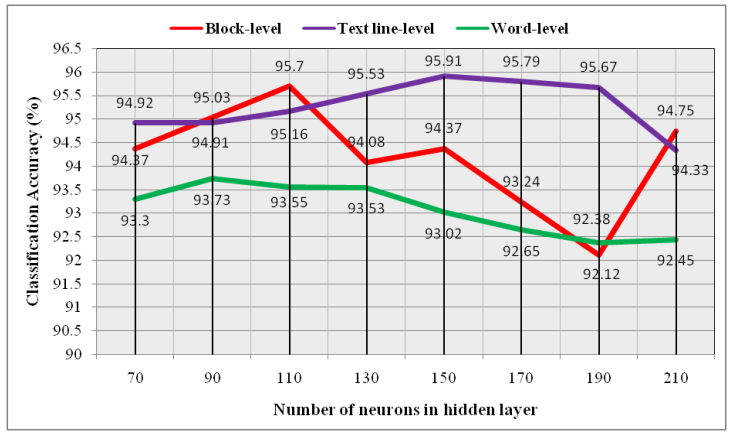

(a)

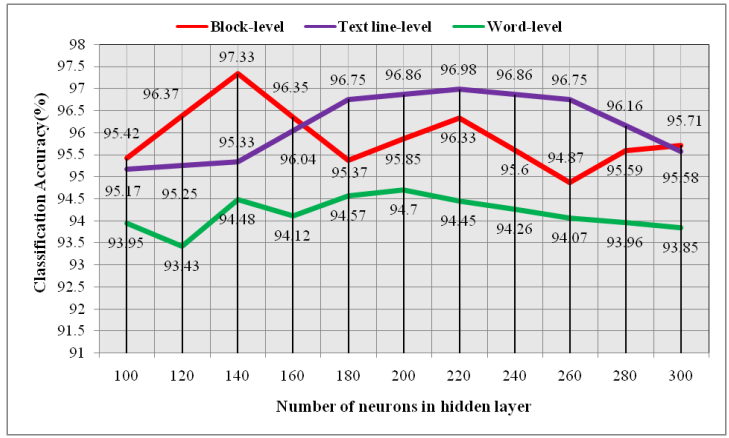

(b)

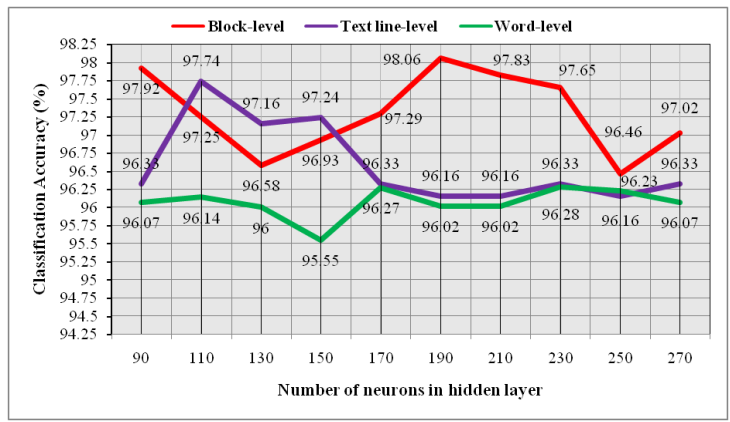

(c)

**Fig. 2: Graphical comparison showing the variation of number of optimal features with the number of neurons in hidden layer of MLP classifier for: (a) DHT algorithm, (b) HOG and (c) MLG transform feature sets on block-level, text line-level, and word-level datasets.**

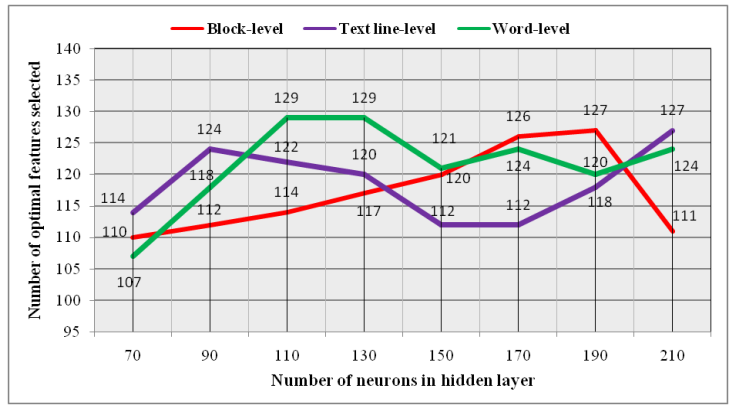

(a)

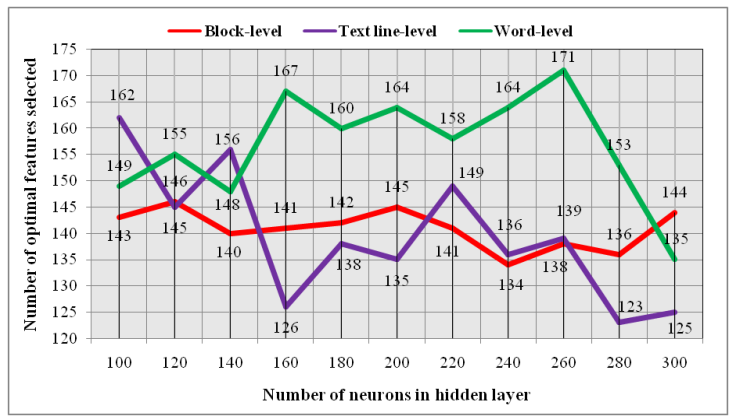

(b)

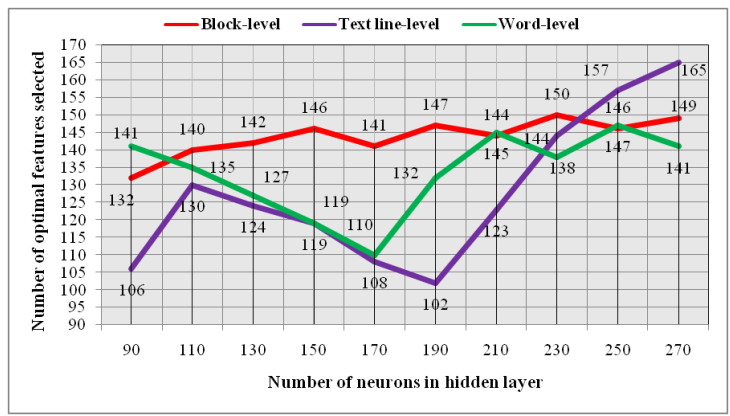

(c)

**Fig. 6:** Graphical comparison showing the variation of script classification accuracy with different values of *k* of *k*-NN classifier for: (a) DHT algorithm, (b) HOG and (c) MLG transform feature sets on block-level, text line-level, and word-level datasets.

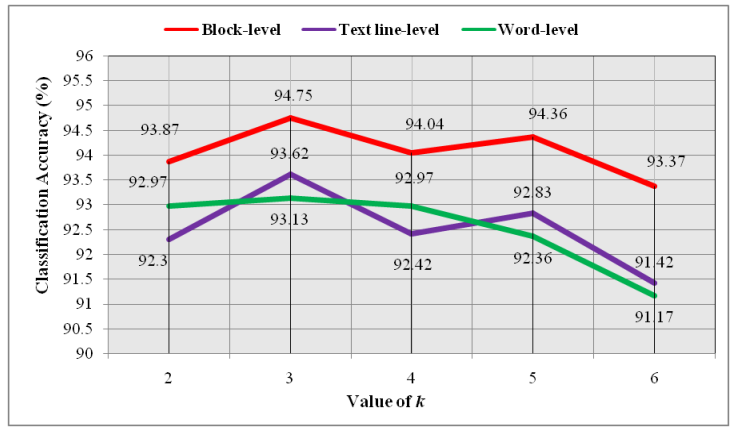

(a)

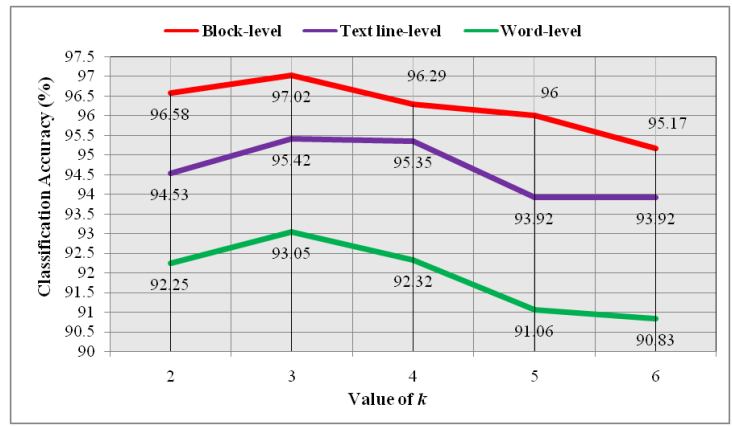

(b)

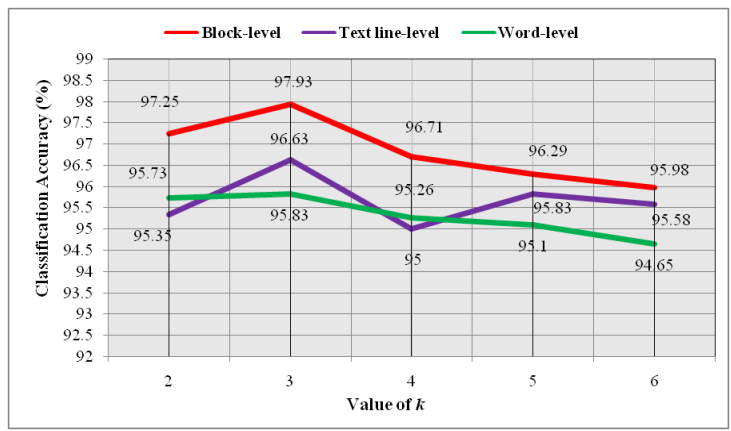

(c)

**Fig. 6: Graphical comparison showing the variation of number of optimal features with different values of *k* of *k*-NN classifier for: (a) DHT algorithm, (b) HOG and (c) MLG transform feature sets on block-level, text line-level, and word-level datasets.**

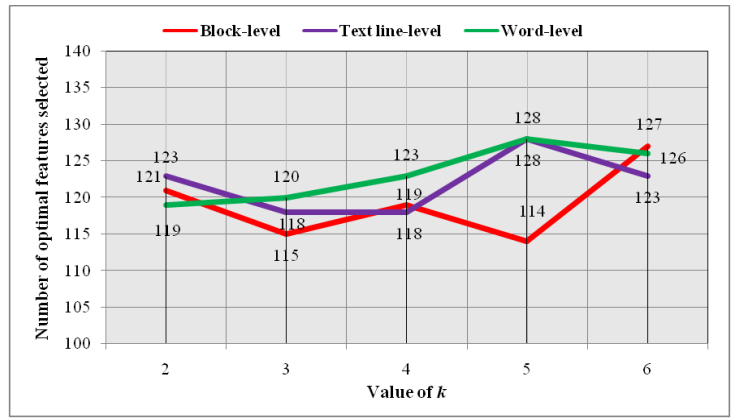

(a)

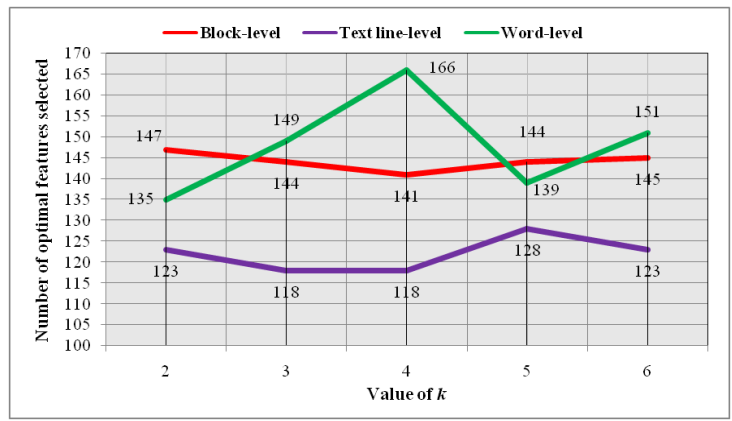

(b)

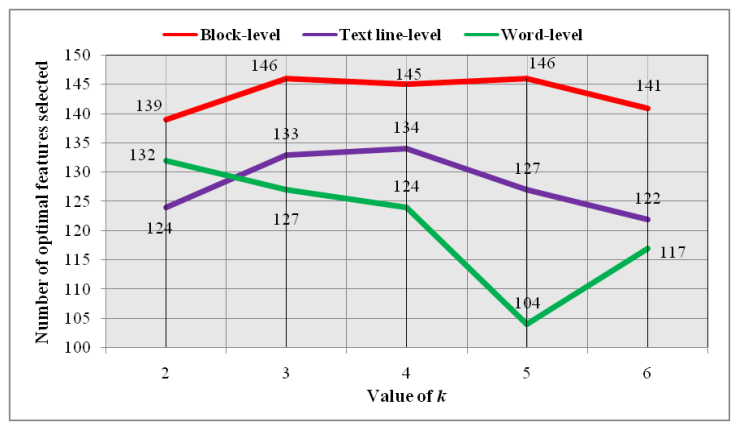

(c)